\documentclass[10pt,twocolumn,letterpaper]{article}

\usepackage[pagenumbers]{cvpr} 

\definecolor{cvprblue}{rgb}{0.21,0.49,0.74}
\usepackage[pagebackref,breaklinks,colorlinks,allcolors=cvprblue]{hyperref}

\usepackage{algorithm}
\usepackage{algorithmic}
\usepackage{graphicx}
\usepackage{booktabs}
\usepackage{multirow}
\usepackage[table]{xcolor}
\usepackage{amsmath,amssymb}
\usepackage{adjustbox}
\usepackage[table]{xcolor}
\definecolor{best}{RGB}{255,210,210}     
\definecolor{second}{RGB}{255,235,200}   
\definecolor{third}{RGB}{255,250,200}    
\newcommand{\best}[1]{\cellcolor{best}{#1}}
\newcommand{\second}[1]{\cellcolor{second}{#1}}
\newcommand{\third}[1]{\cellcolor{third}{#1}}

\def\confName{CVPR}
\def\confYear{2027}

\title{PixIE: Prompted Pixel-Space Low-Light Image Enhancement}
\author{
Ruirui Lin$^{1}$ \quad Jing Han$^{2}$ \quad Guoxi Huang$^{1}$ \quad David Bull$^{1}$ \quad Nantheera Anantrasirichai$^{1}$\\
{\small $^{1}$Visual Information Laboratory}\\
{\small $^{2}$Smart Internet Lab}\\
{\small University of Bristol, United Kingdom}
}

\makeatletter
\g@addto@macro\@maketitle{
    \begin{center}
        \centering
        \vspace{-4mm}
        \includegraphics[width=\textwidth]{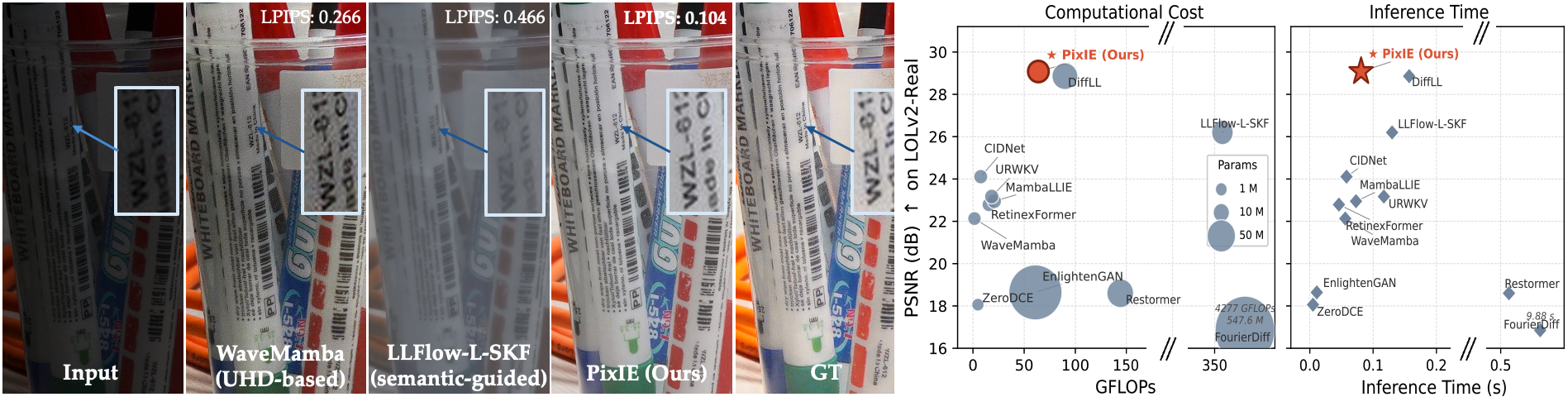}
        \captionof{figure}{\textbf{Pixel-space restoration with spatially continuous FM guidance.} \textbf{Left:} Qualitative comparison on UHD-LL~\cite{UHDFourICLR2023} using models trained on LOLv2-Real~\cite{yang2021sparse}, demonstrating cross-dataset generalization. PixIE suppresses noise and preserves finer textures than UHD-based WaveMamba and semantic-guided LLFlow-L-SKF. \textbf{Right:} PixIE achieves the highest PSNR with competitive GFLOPs and inference time on LOLv2-Real.}
        \label{fig:teaser}
        \vspace{-1mm}
    \end{center}
}
\makeatother

\begin{document}
\maketitle

\begin{abstract}
Low-light images suffer from severe noise, contrast loss, and semantic ambiguity, making enhancement a joint problem of denoising and detail recovery. We propose PixIE, a feed-forward pixel-space LLIE framework semantically prompted by a foundation model (FM). PixIE first performs cross-scale denoising to suppress noise while preserving structure, then refines details using Prompted Pixel Blocks (PPBs), which inject intermediate FM features via spatially continuous modulation (SCMo). For efficient pixel-space attention, we introduce Spatial-Channel Compaction (SCC) to jointly reduce spatial tokens and channel dimensions. We further propose Multi-Receptive-Field Pixel Embedding (MRPE) to provide neighborhood-aware pixel representations before prompting. Experiments demonstrate state-of-the-art performance on standard LLIE benchmarks, achieving the best PSNR, SSIM, and LPIPS on LOLv2-Real, with sharper details and more consistent textures. Code will be publicly available.
\end{abstract}
    
\section{Introduction}
\label{sec:intro}

Low-light image enhancement (LLIE) involves more than brightness adjustment. Severe signal-dependent noise, texture collapse, and color distortion make enhancement a coupled problem of denoising, illumination restoration, and semantically consistent detail recovery. LLIE is also critical for downstream vision tasks such as detection, tracking, and recognition, where poor illumination substantially degrades performance.


Recent LLIE methods based on Convolutional Neural Networks~\cite{lore2017llnet}, Transformers~\cite{xu2022snr,Yan_HVICID}, Mamba models~\cite{zou2024wavemamba}, diffusion models~\cite{hou2023global}, and Retinex-inspired hybrids~\cite{Cai_2023_ICCV,Bai2024RetinexmambaRM} have achieved significant progress. However, data-driven models are limited by their training datasets. 
Under severe underexposure, noise and texture become difficult to distinguish, often causing over-smoothing, color drift, or semantically inconsistent enhancement~\cite{wu2023skf}. This motivates leveraging external semantic priors.

\begin{figure*}[t]
    \vspace{-2mm}
    \centering
    \includegraphics[width=\textwidth]{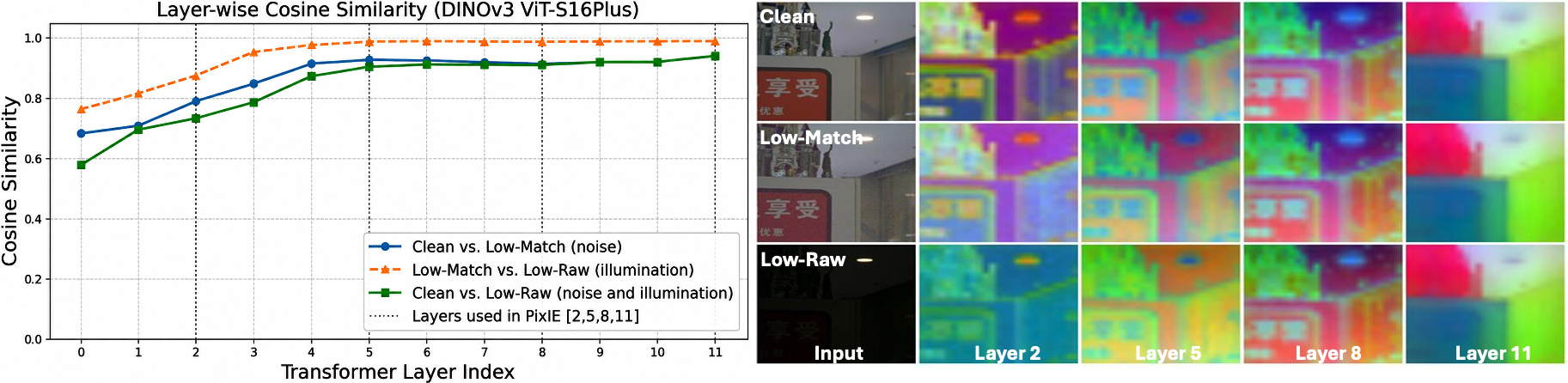}

    \caption{\small DINOv3 ViT-S/16Plus features under low-light degradation.
\textbf{Left:} Mean token-wise cosine similarity across the 12 transformer layers for three comparisons: Clean vs.\ Low-Match (noise), Low-Match vs.\ Low-Raw (illumination), and Clean vs.\ Low-Raw (noise and illumination). Similarity increases with depth, indicating progressively more stable representations; dotted vertical lines mark the layers $\{2,5,8,11\}$ used for PixIE conditioning.
\textbf{Right:} PCA visualizations at selected layers show the same trend.}
    \label{fig:showDINO}
        \vspace{-4mm}
\end{figure*}

Several restoration methods have explored foundation-model (FM) features from CLIP~\cite{Radford2021LearningTV}, Segment Anything Model (SAM)~\cite{ravi2024sam2}, and DINO~\cite{simeoni2025dinov3} as guidance signals~\cite{jin2023let,qi2024spire,Luo2023ControllingVM}. Our analysis shows that DINOv3~\cite{simeoni2025dinov3} provides robust dense features under low-light degradation, helping reduce semantic ambiguity in noisy regions, as shown in~\cref{fig:showDINO} and confirmed in~\cref{tab:ablation}. Existing methods exploit semantic priors through various feature integration strategies~\cite{wu2023skf,Luo2023ControllingVM,xu2024boosting}, but effectively aligning such guidance with pixel-level restoration remains challenging. Under spatially varying illumination and noise, insufficiently fine-grained guidance can lead to cross-region inconsistencies and over-smoothed local structures, which are often overlooked as enhancement methods primarily emphasize global illumination correction.


While pixel-space modeling better preserves high-frequency details, it also introduces two challenges. First, naive pixel-space attention is costly at high resolutions, making pixel-space enhancement underexplored~\cite{SimpleDiff_pix_23,HourglassDiT}. Second, many pixel-space architectures rely on point-wise ($1\times1$) projections~\cite{yu2025pixeldit}, which have no spatial receptive field. Under severe low-light noise, relying only on single-pixel statistics makes it difficult to separate structure from noise~\cite{wei2021physics,Foi_2008_PG,Multiscale_Guo}, motivating local context aggregation before semantic prompting.

In this paper, we propose \textbf{PixIE}, an end-to-end \textbf{Pix}el-space LL\textbf{IE} framework prompted by a frozen DINOv3 encoder. PixIE performs feed-forward restoration on pixel-aligned features at full resolution. To the best of our knowledge, PixIE is the first LLIE method to introduce spatially continuous modulation (SCMo) for integrating FM guidance into pixel-aligned restoration.
As shown in~\cref{fig:teaser}, PixIE suppresses low-light noise while preserving fine structures more effectively than other LLIE methods. It also achieves a favorable performance–complexity trade-off. PixIE first applies a semantics-free cross-scale denoising stream to suppress severe noise while preserving structure, then uses Multi-Receptive-Field Pixel Embedding (MRPE) to equip pixels with neighborhood context before prompting. The resulting features are refined by Prompted Pixel Blocks (PPBs), which use prompted features to predict modulation directly on the FM token grid and upsample it into a spatially continuous modulation field for pixel-level restoration. This enables semantically guided detail recovery without hard patch boundaries, which are particularly visible under spatially varying illumination correction as validated in~\cref{sec:abl}. Finally, Spatial-Channel Compaction (SCC) makes pixel-space attention affordable across scales by jointly reducing the spatial token grid and channel dimension.

In summary, our main contributions are:
\begin{itemize}
    \item We propose PixIE, a novel feed-forward pixel-space LLIE framework that denoises before prompting and guides pixel-aligned features using a foundation model (FM) guidance.
    \item We introduce PPBs with spatially continuous modulation (SCMo), which predicts modulation directly on the FM token grid and upsamples it into a continuous field.
    \item We propose SCC to make pixel-space attention efficient by reducing both spatial tokens and channel dimensions.
    \item We introduce MRPE to aggregate multi-receptive-field neighbourhood context before prompting.
\end{itemize}
\section{Related Work}
\subsection{Low-light Image Enhancement (LLIE).} 
Traditional LLIE methods typically rely on hand-crafted priors and manual parameter tuning, and often suffer from over-amplification and halo artifacts. Representative approaches include histogram equalization~\cite{ibrahim2007brightness} and Retinex-based methods~\cite{Land1977TheRT}, which decompose an image into illumination and reflectance components. 
Recent LLIE methods span CNNs~\cite{Lv2018MBLLEN,lore2017llnet}, Transformers~\cite{Cai_2023_ICCV, xu2022snr}, Retinex-inspired networks~\cite{Cai_2023_ICCV,Bai2024RetinexmambaRM}, diffusion models~\cite{jiang2023low,Jiang_2024_ECCV}, and Mamba architectures~\cite{zou2024wavemamba,Bai2024RetinexmambaRM,huang2026bayesian}. Some further exploit semantic priors in feature space~\cite{zheng2022semantic,wu2023skf} or alternative color representations~\cite{Yan_HVICID}. Despite these advances, most methods emphasize illumination correction and denoising, often sacrificing fine textures in challenging low-light scenes. In contrast, PixIE achieves detail-preserving enhancement while remaining robust to noise.

\begin{figure*}[t]
\vspace{-2mm}
    \centering
    \includegraphics[width=\linewidth]{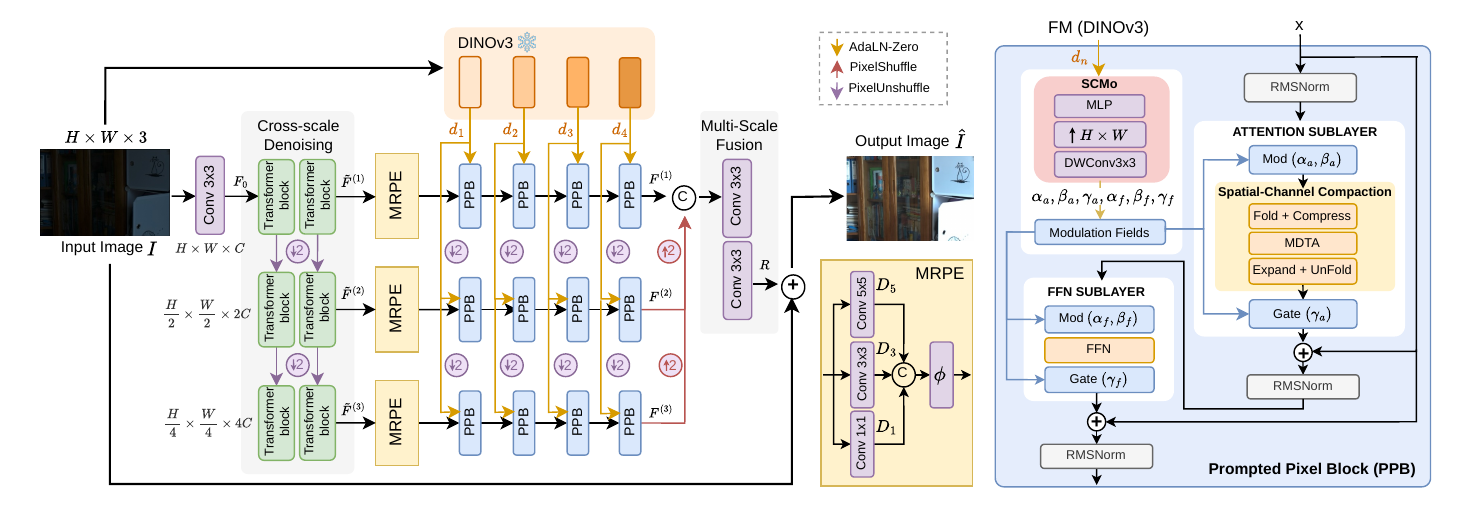}
    \caption{Overview of PixIE. Cross-scale denoising suppresses noise from fine to coarse scales, followed by MRPE for local context aggregation and PPBs for FM-guided pixel-space refinement. Multi-scale fusion combines the refined features to predict a residual correction, $\hat{I}=I+R$.}
    \label{fig:arch}
    \vspace{-4mm}
\end{figure*}

\subsection{Pixel-Space Generative Models.}
Latent-space pipelines~\cite{Peebles_23,Rombach2021HighResolutionIS}, widely adopted in modern generative models, improve computational efficiency by compressing images into low-dimensional representations. However, their patch-based tokenization entangles adjacent pixels, limiting fine-grained detail recovery and hindering high-fidelity generation. 
To address this, pixel-space modeling~\cite{SimpleDiff_pix_23} captures dense per-pixel interactions, preserving high-frequency details. Recent generative models increasingly adopt pixel-space diffusion~\cite{chen2025dip,yu2025pixeldit,zheng2025diffusion} to avoid artifacts introduced by variational autoencoder compression~\cite{VAE} and patch-based tokenization. For image restoration, pixel-space processing is important for texture fidelity but presents a trade-off: global attention is computationally expensive at high resolutions~\cite{Zamir2021Restormer}, while standard $1\times1$ projections lack sufficient receptive fields to distinguish signal from noise~\cite{Chang_20ECCV}. PixIE adopts a pixel-space framework, \ie operating on pixel-aligned features rather than compressed image latents, but performs feed-forward restoration rather than generative modeling. SCC enables efficient attention while retaining pixel-aligned features at each scale.

\subsection{Foundation Models for Image Restoration.}

Recent foundation models (FMs), including CLIP, DINO, and SAM~\cite{Radford2021LearningTV,simeoni2025dinov3,ravi2024sam2}, have shown strong robustness to image degradations and are increasingly used as semantic priors for image restoration. CLIP features provide linguistic-visual guidance for color and style stabilization~\cite{Luo2023ControllingVM,Xue_CLIPLL,wang2022clipiqa}, while SAM supplies object-level masks for region-aware restoration~\cite{Zhang2024DistillingSP,li2024sam}. However, integrating these priors into a pixel-level restoration pipeline remains challenging and underexplored. Existing methods exploit semantic priors through semantic-guided learning~\cite{zheng2022semantic,wu2023skf}, attention-based fusion~\cite{Luo2023ControllingVM,lin2023multi}, or spatially varying enhancement~\cite{xu2024boosting}, leaving a semantic-spatial gap that limits pixel-level restoration. Spatial feature transform~\cite{wang2018sftgan} further enables spatially adaptive modulation by upsampling the conditioning features and predicting per-pixel parameters at feature resolution. In contrast, our Prompted Pixel Block (PPB) uses FM features with SCMo predicting modulation directly on the FM token grid and upsampling it into a spatially continuous field for pixel-level restoration.
\section{Methodology}

As illustrated in~\cref{fig:arch}, PixIE consists of three components:
(i) \emph{Cross-Scale Denoising}, which suppresses noise and preserves coarse structure using efficient Transformer blocks with fine-to-coarse feature propagation;
(ii) \emph{Pixel-Space Enhancement}, which enriches local context via Multi-Receptive-Field Pixel Embedding (MRPE) and enhances in pixel-space using FM guidance via Prompted Pixel Blocks (PPB);
and (iii) \emph{Multi-Scale Fusion}, which aggregates the refined features from all scales to predict a residual enhancement.

Given a low-light input $I\in\mathbb{R}^{B\times 3\times H\times W}$, we first extract features $F_0\in\mathbb{R}^{B\times C\times H\times W}$ via a 3$\times 3$ convolution, expanding the channel capacity for subsequent transformer blocks. $B$ is the batch size, $(H,W)$ is the spatial resolution, and $C$ is the channel dimension.

We then construct a three-scale pyramid $s\in\{1,2,3\}$, corresponding to resolutions of $(H,W,C)$, $(H/2,W/2,2C)$, and $(H/4,W/4,4C)$, respectively. The denoising stream operates fine-to-coarse and produces denoised features $\tilde{F}^{(s)}$.
These are then processed by MRPE and a stack of $N$ PPBs (we set $N=4$) to obtain semantically refined features $F^{(s)}$. Specifically, DINOv3 is frozen and executed once per input. We extract intermediate-layer tokens and resize the token grids to provide semantic prompts at all scales.
Cross-scale propagation is restricted to the denoising stream (see~\cref{sec:denoising}): only denoised features $\tilde{F}^{(s)}$ are forwarded to the next coarser stage, while the refined PPB outputs, $F^{(s)}$, are used only for final fusion and are not fed back into denoising blocks to avoid propagating semantically modulated pixel-space features across scales.
Finally, we upsample $F^{(2)}$ and $F^{(3)}$ via PixelShuffle, concatenate them with $F^{(1)}$, and apply a simple Multi-Scale Fusion with two $3\times3$ convolutions to predict a residual $R$, yielding the final output $\hat{I}=I+R$.


\subsection{Cross-Scale Denoising} 
\label{sec:denoising}
The cross-scale denoising module produces a noise-suppressed, structurally preserved representation before semantic prompting.
Under low-light conditions, severe noise and degradation corrupt local information, making alignment with high-level semantic priors unreliable and potentially unstable. To address this, we first apply the denoiser, keeping the denoising stream semantics-free and injecting DINO conditioning only in subsequent PPBs, where cleaner features allow more reliable spatial alignment.~\cref{sec:abl}~\cref{tab:ablation} further supports this. Moreover, we propagate only denoised features across scales to avoid feeding semantically modulated activations back into the denoiser, which might introduce distribution shifts across scales.

We adopt a Transformer-based denoiser, as the self-attention mechanism can be formulated as a non-local operation in the embedded Gaussian form \cite{Lin:muliti:2025}. For efficiency, we employ Restormer-style blocks, each composed of Multi-Dconv Head Transposed Attention (MDTA) followed by a Gated-Dconv Feed-Forward Network (GDFN). 
Since MDTA attends in the channel domain, its computational cost scales as $\mathcal{O}(HW\cdot C^{2})$, \ie, linearly with the number of pixels for a fixed channel width.

Rather than processing scales independently, we propagate features fine-to-coarse: each transformer block's output is downsampled and fused into the corresponding block at the next coarser scale. This allows fine-scale branches to pass explicit structural cues to coarser representations, where a larger effective receptive field facilitates suppression of spatially correlated noise. At each scale $s$, the denoising stream contains two sequential transformer blocks $\mathcal{T}^{(s)}_1$ and $\mathcal{T}^{(s)}_2$. Let $\tilde{F}^{(s)}_0$ denote the input features at scale $s$. The blocks process features sequentially:
\begin{equation}
\tilde{F}^{(s)}_1 = \mathcal{T}^{(s)}_1\!\left(\tilde{F}^{(s)}_0\right), 
\qquad
\tilde{F}^{(s)}_2 = \mathcal{T}^{(s)}_2\!\left(\tilde{F}^{(s)}_1\right).
\end{equation}
where $\tilde{F}^{(s)} = \tilde{F}^{(s)}_2$ denotes the denoised feature at scale $s$.

For scales $s>1$, the output of each block at the finer scale is downsampled via PixelUnshuffle and fused with the corresponding block output at the current scale via concatenation:
%
\begin{equation}
\begin{aligned}
\tilde{F}^{(s)}_k \leftarrow 
W^{(s)}_k\!\Big(
&\big[\tilde{F}^{(s)}_k\,;\,
\mathrm{Down}\!\left(\tilde{F}^{(s-1)}_k\right)\big]
\Big), \\
& k \in \{1,2\},\quad s \in \{2,3\}.
\end{aligned}
\end{equation}
where $[\,\cdot\,;\,\cdot\,]$ denotes channel-wise concatenation, $W^{(s)}_k$ is a learned channel projection, and $\mathrm{Down}(\cdot)$ denotes PixelUnshuffle downsampling. For $s>1$, $\tilde{F}^{(s)}_0=\mathrm{Down}(\tilde{F}^{(s-1)}_1)$, and the fused $\tilde{F}^{(s)}_1$ from Eq.~(2) is passed to $\mathcal{T}^{(s)}_2$. The denoised features $\tilde{F}^{(s)}$ are then passed to MRPE and the PPB stack at scale $s$.

\subsection{Pixel-Space Enhancement}
\begin{figure}
\vspace{-2mm}
    \centering
    \includegraphics[
        width=\columnwidth,
        trim=20 4 25 4,
        clip
    ]{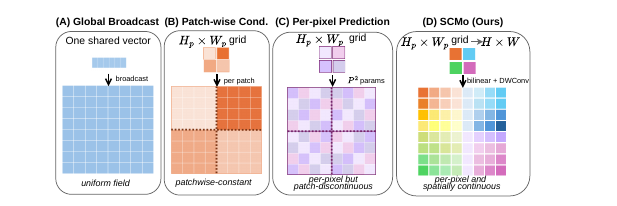}
    \caption{Comparison of modulation strategies. (A) Global broadcast. (B) Patch-wise conditioning produces boundary seams. (C) Per-pixel prediction of $P^2$ modulation parameters per patch causes discontinuities. (D) SCMo predicts modulation on the FM token grid and upsamples it via bilinear interpolation and DWConv, producing spatially continuous, boundary-free modulation.}
\vspace{-4mm}
    \label{fig:mod}
\end{figure}

\paragraph{\textbf{Multi-Receptive-Field Pixel Embedding (MRPE).}}
We introduce MRPE as a simple local context aggregation module placed before our pixel-level restoration blocks. MRPE computes three parallel embeddings with kernel sizes $k\in\{1,3,5\}$, capturing point-wise intensity information with local structural context.
We allocate the embedding dimension $C$ across the three branches to emphasize contextual cues: specifically, we assign $25\%$ of the channels to the $1\times1$ branch and distribute the remaining $75\%$ equally to the $3\times3$ and $5\times5$ branches, \ie, $(D_{1},D_{3},D_{5})=(0.25D,\,0.375D,\,0.375D)$. This design is motivated by low-light degradation, where single-pixel intensities are often dominated by signal-dependent shot noise (whose variance increases with pixel intensity). We therefore allocate fewer channels to the point-wise branch and more to larger receptive fields for robust, neighborhood-aware denoising, as supported in~\cref{sec:abl}.
The resulting features are concatenated and linearly mixed:
\[
e=\phi\!\left([f_{1}(\tilde{F}^{(s)});f_{3}(\tilde{F}^{(s)});f_{5}(\tilde{F}^{(s)})]\right),
\]
where $[\cdot;\cdot]$ denotes channel-wise concatenation,  $f_k(\cdot)$ denotes a $k\times k$ convolution and $\phi(\cdot)$ is a linear projection.
By injecting multi-scale spatial context early, MRPE produces a denoising-friendly embedding $e$ for the subsequent PPB stack.

\paragraph{\textbf{Prompted Pixel Block (PPB).}}
The proposed PPB, shown in~\cref{fig:arch} (blue panel), is a pixel-level residual block that injects semantic guidance from a frozen foundation model (FM), \eg DINOv3, via the proposed Spatially Continuous Modulation (SCMo). 
Rather than using only the final-layer representation of the FM, which is highly invariant but may lose fine spatial details, we extract intermediate patch tokens from $L$ selected layers of the frozen FM encoder. For DINOv3, we set $L{=}4$ and take layers $\{2,5,8,11\}$, producing $L$ token grids $\{d_n\}_{n=1}^{L}$, where $d_n\in\mathbb{R}^{B\times D_d\times H_p\times W_p}$, and $D_d$ denotes the dimension of the FM embedding.
For each pyramid scale $s$, we resize the token grids to $H_p^{(s)}\!\times W_p^{(s)}$ and use them to condition a stack of $N$ PPBs, with each PPB receiving the corresponding token grid.

Given the input feature $d_x\in\mathbb{R}^{B\times C\times H\times W}$, the corresponding FM token grid $d_n\in\mathbb{R}^{B\times D_d\times H_p\times W_p}$ is used to predict AdaLN-Zero parameters with a lightweight MLP, where $n\in\{1,..,L\}$, $H_p=H/P$ and $W_p=W/P$, with $P$ denoting the patch size:
\begin{equation}
m_n=\mathrm{MLP}(d_n)\in\mathbb{R}^{B\times 6C\times H_p\times W_p}.
\end{equation}
In our SCMo, we predict modulation on the token grid $m_n\in\mathbb{R}^{B\times 6C\times H_p\times W_p}$, and upsample it to a full-resolution, spatially smooth parameter field $\hat m_n\in\mathbb{R}^{B\times 6C\times H\times W}$ via $\hat m_n=\mathrm{DWConv}_{3\times3}(\mathrm{Up}(m_n))$, where $\mathrm{Up}(\cdot)$ upsamples via bilinear interpolation from $H_p\times W_p$ to $H\times W$, and $\mathrm{DWConv}(\cdot)$ denotes depth-wise convolution. This yields a spatially continuous modulation by blending adjacent tokens near patch boundaries as shown in~\cref{fig:mod}(D).

This differs from PixelDiT~\cite{yu2025pixeldit} (see~\cref{fig:mod}(C)), which predicts $P^2$ modulation parameter sets per token independently within each patch at $\mathcal{O}(D_d \cdot P^2 C)$ MLP cost, resulting in per-pixel but patch-discontinuous modulation with hard discontinuities. Our approach predicts a single parameter set per token 
($\mathcal{O}(D_d \cdot C)$ MLP cost), recovering full-resolution modulation via interpolation rather than direct per-pixel prediction, achieving $P^2{\times}$ 
fewer parameters in the conditioning MLP ($256{\times}$ at $P{=}16$) while producing a smoother, continuous modulation field. In low-light enhancement, where spatially varying exposure corrections are large, patch-boundary discontinuities are particularly visible (see~\cref{sec:abl}~\cref{fig:abl}). 

We split $\hat{m}_n$ channel-wise into
$\{\alpha_a,\beta_a,\gamma_a,\alpha_f,\beta_f,\gamma_f\}_n \\ \in\mathbb{R}^{B\times C\times H\times W}$,
corresponding to scale, shift, and gating for the attention ($\cdot_a$) and feed-forward ($\cdot_f$) sublayers.
Let $\mathrm{RMSNorm}(\cdot)$ denote Root Mean Square Normalization and
$\mathrm{Mod}(u;\alpha,\beta)=\alpha\odot u + \beta$.
The attention (Att) and the feed-forward network (FFN) updates are
\begin{align}
x &\leftarrow x + \gamma_a \odot \mathrm{Attn}\!\left(\mathrm{Mod}(\mathrm{RMSNorm}(x);\alpha_a,\beta_a)\right),\\
x &\leftarrow x + \gamma_f \odot \mathrm{FFN}\!\left(\mathrm{Mod}(\mathrm{RMSNorm}(x);\alpha_f,\beta_f)\right).
\end{align}

\paragraph{\textbf{Spatial-Channel Compaction (SCC).}}
To balance efficiency and visual quality, the SCC (yellow panel inside blue panel in~\cref{fig:arch}) is applied only to the attention branch, while the FFN, responsible for per-pixel feature transformation, operates on the full-resolution feature map to preserve and restore local details.

Although MDTA avoids quadratic \emph{spatial} attention, its \emph{channel} attention scales as $\mathcal{O}(HW\cdot C^{2})$, which becomes expensive in multi-scale design as $C$ increases at coarser levels.
To reduce this cost, we compact features in both spatial and channel dimensions before attention.
SCC serves only the attention branch of each PPB and differs from PixelDiT~\cite{yu2025pixeldit} in three aspects: (i) compression to $D_{\mathrm{attn}}$, (ii) MDTA instead of spatial attention (see~\cref{sec:abl}), and (iii) restoration across multi-scales, with compacted attention cost itself independent of original channel width $C$. 
Given $x\in\mathbb{R}^{B\times C\times H\times W}$, we first fold it into non-overlapping $P\times P$ patches,
$x_{\mathrm{fold}}\in\mathbb{R}^{B\times (CP^{2})\times H_p\times W_p}$, where $H_p=H/P$ and $W_p=W/P$.
We then apply a $1\times1$ projection to compress the inflated channels to an attention dimension
$x_{\mathrm{c}}\in\mathbb{R}^{B\times D_{\mathrm{attn}}\times H_p\times W_p}$ (we use $D_{\mathrm{attn}}=16$), perform MDTA on $x_{\mathrm{c}}$, and map back via a $1\times1$ expansion followed by patch unfolding.
The resulting attention cost is $\mathcal{O}(H_pW_p\cdot D_{\mathrm{attn}}^{2})$, since MDTA is performed on the compact representation $x_{\mathrm{c}}$.
Compared to applying MDTA directly on $x$ with cost $\mathcal{O}(HW\cdot C^{2})$, spatial folding reduces the number of locations by a factor of $P^{2}$ ($HW\rightarrow H_pW_p$), while channel bottlenecking replaces the quadratic channel term $C^{2}$ with $D_{\mathrm{attn}}^{2}$ (with $D_{\mathrm{attn}}\ll CP^{2}$ in the folded space), making the attention cost insensitive to the increasing $C$ at coarser scales.

\begin{table*}[!t]
\centering
\caption{Quantitative comparison on LOLv1, LOLv2 (Real and Synthetic), and LSRW. Cells highlighted in \best{red}, \second{orange}, and \third{yellow} denote the best, second-best, and third-best results, respectively.}

\footnotesize
\small
\resizebox{\textwidth}{!}{%
\begin{tabular}{l|ccc|ccc|ccc|ccc}
\toprule
\multirow{2}{*}{Methods}
& \multicolumn{3}{c|}{LOLv1}
& \multicolumn{3}{c|}{LOLv2-Real}
& \multicolumn{3}{c|}{LOLv2-Synthetic}
& \multicolumn{3}{c}{LSRW} \\
\cmidrule(lr){2-4}\cmidrule(lr){5-7}\cmidrule(lr){8-10}\cmidrule(lr){11-13}
& PSNR$\uparrow$ & SSIM$\uparrow$ & LPIPS$\downarrow$
& PSNR$\uparrow$ & SSIM$\uparrow$ & LPIPS$\downarrow$
& PSNR$\uparrow$ & SSIM$\uparrow$ & LPIPS$\downarrow$
& PSNR$\uparrow$ & SSIM$\uparrow$ & LPIPS$\downarrow$ \\
\midrule
ZeroDCE~\cite{Zero-DCE}             & 14.86 & 0.562 & 0.372 & 18.06 & 0.580 & 0.352 & 17.71 & 0.815 & 0.169 & 15.87 & 0.443 & 0.411 \\
EnlightenGAN~\cite{jiang2021enlightengan}   & 17.48 & 0.652 & 0.322 & 18.64 & 0.677 & 0.309 & 16.57 & 0.734 & 0.220 & 17.11 & 0.463 & 0.406 \\
Restormer~\cite{Zamir2021Restormer}     & 22.43 & 0.823 & 0.147 & 18.60 & 0.789 & 0.232 & 21.41 & 0.830 & 0.144 & 16.30 & 0.453 & 0.427 \\
RetinexFormer~\cite{Cai_2023_ICCV} & 25.16 & 0.845 & 0.131 & 22.79 & 0.840 & 0.171 & 25.67 & 0.930 & 0.059 & 17.77  & 0.518 &  \third{0.317} \\
DiffLL~\cite{jiang2023low}                         & 26.34 & 0.845 & 0.217 & \second{28.86} & 0.876 & 0.207 & \second{28.65} & 0.940 & 0.082 & \third{19.28} & \second{0.552} & 0.350 \\
LLFlow-L-SKF~\cite{wu2023skf}                          & 25.13 & 0.872 & 0.117 & \third{26.20} & \third{0.888} & 0.137 & 24.81 & 0.919 & 0.067 & 16.26 & 0.457 & 0.390 \\
FourierDiff~\cite{lv2024fourier} & 17.56 &0.611  &0.287 &16.85 & 0.607 & 0.294 &14.20 & 0.651 & 0.284  &15.64  &0.458  &0.327  \\
WaveMamba~\cite{zou2024wavemamba}                          & \third{26.54} & \second{0.883} & 0.106 & 22.13 & \second{0.890} & 0.162 & \third{26.51} & \best{0.957} & 0.065 & \second{19.51} & \best{0.569} & 0.300 \\
MambaLLIE~\cite{MambaLLIE} & 21.54 & 0.816 & 0.154 &22.95 & 0.847 & 0.171 & 25.87 & 0.940 & 0.053  &17.02  &0.502  & 0.346 \\

URWKV~\cite{xu2025urwkv} & 23.54  &0.857  &\third{0.104} &23.11  & 0.874  & \third{0.132} & 26.63 & \third{0.944}  & \best{0.031}  & 17.17 & 0.530 & 0.328  \\
CIDNet~\cite{Yan_HVICID}      & \best{28.20} & \best{0.889} & \second{0.079} & 24.11 & 0.871 & \second{0.108} & 25.71 & 0.942 & \third{0.045} & 18.07  & \third{0.532} & \second{0.286} \\

\midrule
PixIE (Ours)
& \second{27.07} & \third{0.876} & \best{0.073}
& \best{29.08} & \best{0.902} & \best{0.089}
& \best{29.10} & \second{0.950} & \second{0.042}
& \best{19.87} & \third{0.532} & \best{0.271} \\
\bottomrule
\end{tabular}%
}
\label{tab:llie_main}
\end{table*}

\begin{figure*}[t]
    \centering

    \includegraphics[width=\linewidth]{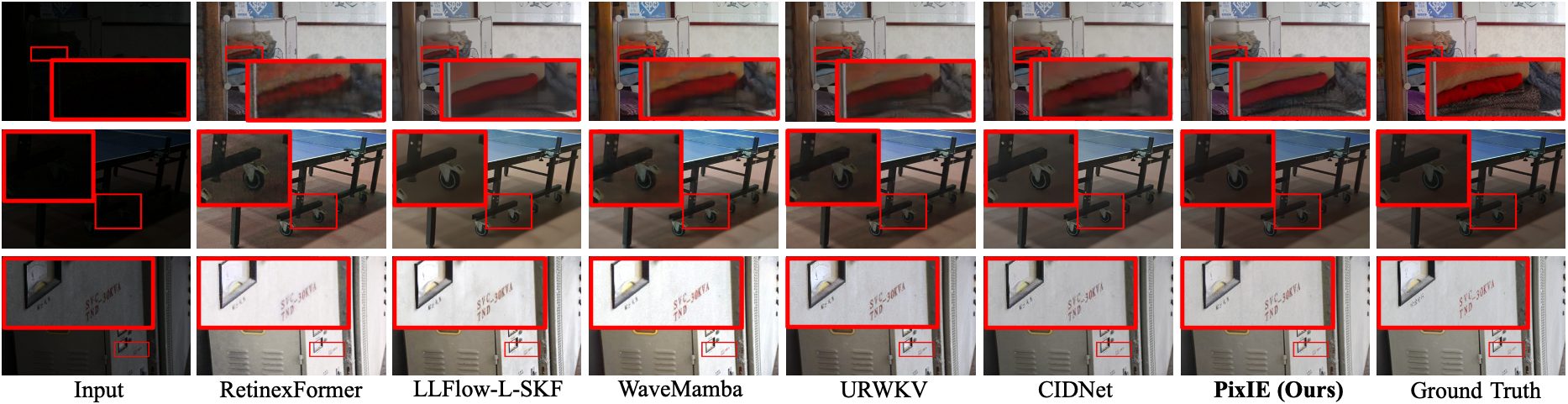}
    \caption{Qualitative comparison of PixIE with recent state-of-the-art methods on LOLv1, LOLv2-Real, and LSRW test sets.}
    \label{fig:vis_results}
        \vspace{-4mm}
\end{figure*}

\begin{figure}[t]
    \centering

    \includegraphics[width=\linewidth]{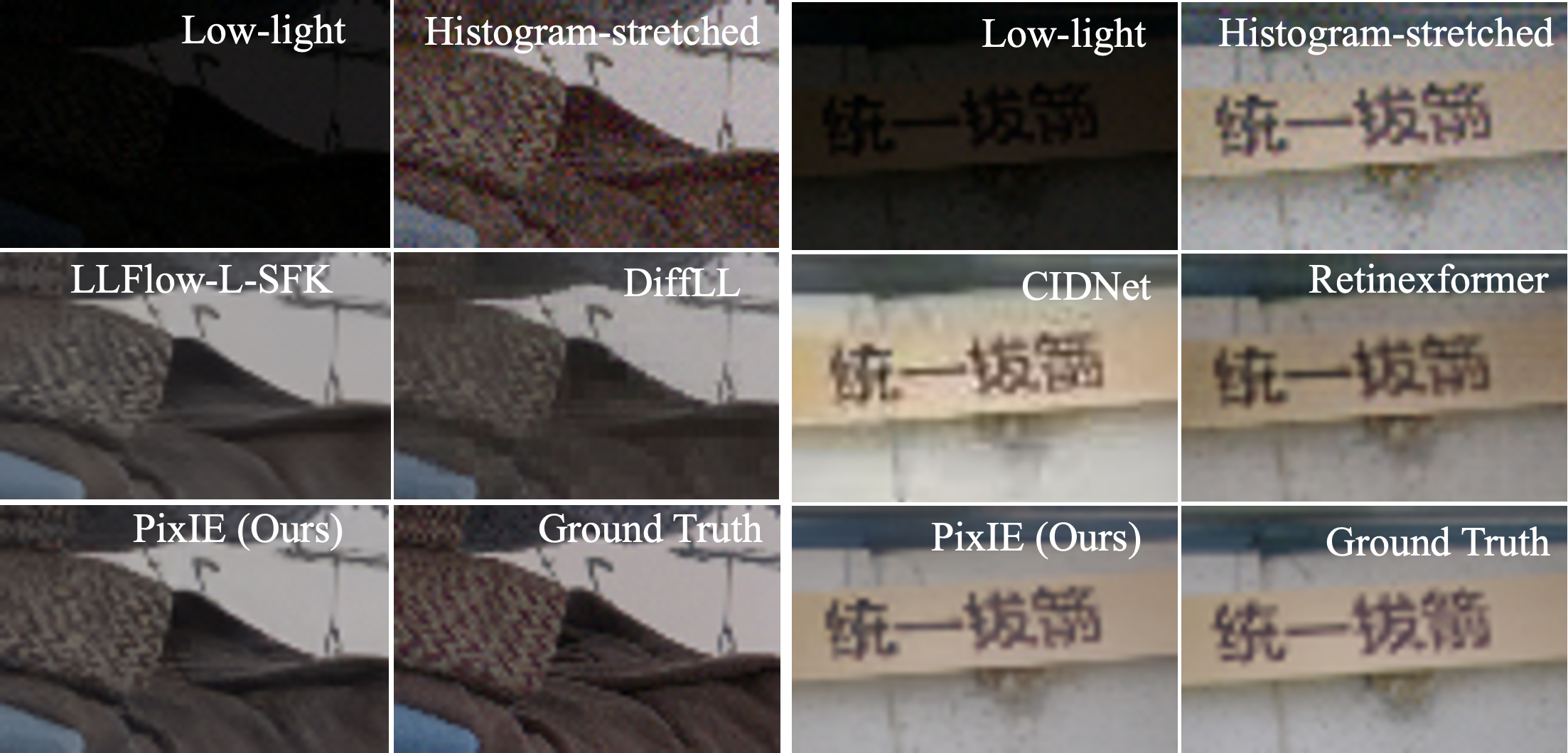}
    \caption{Comparison on  textural areas of LOLv1 and LOLv2-Real. Low-light input is histogram-matched to ground truth for noise visualization.}
    \label{fig:LoLv1v2}
        \vspace{-4mm}
\end{figure}

\begin{table*}[t]
\caption{Quantitative comparison on five unpaired datasets: LIME, VV, DICM, NPE, and MEF. Cells highlighted in \best{red}, \second{orange}, and \third{yellow} denote the best, second-best, and third-best results, respectively.}
\centering
\small
\setlength{\tabcolsep}{2.2pt}
\renewcommand{\arraystretch}{1.08}
\resizebox{\textwidth}{!}{%
\begin{tabular}{l|ccc|ccc|ccc|ccc|ccc}
\toprule
\multirow{2}{*}{Methods}
& \multicolumn{3}{c|}{LIME}& \multicolumn{3}{c|}{VV}& \multicolumn{3}{c|}{DICM}& \multicolumn{3}{c|}{NPE}& \multicolumn{3}{c}{MEF} \\
\cmidrule(lr){2-4}\cmidrule(lr){5-7}\cmidrule(lr){8-10}\cmidrule(lr){11-13}\cmidrule(lr){14-16}
& NIQE$\downarrow$ & MUSIQ$\uparrow$ & CLIP-IQA$\uparrow$
& NIQE$\downarrow$ & MUSIQ$\uparrow$ & CLIP-IQA$\uparrow$
& NIQE$\downarrow$ & MUSIQ$\uparrow$ & CLIP-IQA$\uparrow$
& NIQE$\downarrow$ & MUSIQ$\uparrow$ & CLIP-IQA$\uparrow$
& NIQE$\downarrow$ & MUSIQ$\uparrow$ & CLIP-IQA$\uparrow$ \\
\midrule

ZeroDCE~\cite{Zero-DCE}& \second{3.76} & \second{62.15} & \third{0.454}& 2.96 & \second{59.96} &\second{0.350}& 4.00 & \third{58.77} & 0.513& \third{3.95} & 67.65 & \second{0.458}& \second{3.29} & 64.31 & \second{0.514} \\
EnlightenGAN~\cite{jiang2021enlightengan}& 3.92 & 47.92 & 0.390& 3.01 & 38.47 & 0.290& \third{3.53} & 38.87 & 0.457& 4.01 & 46.76 & 0.364& 3.48 & 49.00 & 0.483 \\
Restormer~\cite{Zamir2021Restormer}& 4.16 & 61.48 & 0.446& 3.26 & 52.04 & \best{0.470}& 4.34 & 56.52 & \third{0.514}&4.08 & 65.70 & \best{0.473}& 4.16 & 60.60 & 0.447 \\
RetinexFormer~\cite{Cai_2023_ICCV}& 4.03 & 56.92 & 0.433& \second{2.82} & 51.40 & 0.259& \best{3.37} & 51.70 & 0.444&4.13 & 58.60 & 0.287& 3.44 & 58.64 & 0.478 \\
DiffLL~\cite{jiang2023low}& \third{3.78} & 50.47 & 0.254& 3.51 & 43.43 & 0.264& 3.85 & 49.87 & 0.211& 4.31 & 55.46 & 0.206& 3.66 & 53.28 & 0.200 \\
LLFlow-L-SFK~\cite{wu2023skf}& 5.26 & \best{63.13} & 0.296& 5.48 & 51.27 & 0.296& 5.53 & 58.65 & 0.304& 6.92 & 64.84 & 0.297& 6.43 & 61.95 & 0.265 \\
WaveMamba~\cite{zou2024wavemamba}& 4.07 & 60.92 & 0.428& 3.05 & 53.37 & 0.306& 3.88 & \second{59.14} & 0.467& \best{3.60} & 67.48 & 0.410& 3.62 & \second{64.97} & 0.479 \\
URWKV~\cite{xu2025urwkv}& \best{3.73} & 60.90 & 0.428& 3.45 & \best{60.13} & 0.328& \second{3.52} & 58.45 & \second{0.532}& 3.99 & \second{69.34} & 0.397& \best{3.22} & 63.84 & 0.489 \\
CIDNet~\cite{Yan_HVICID}& 4.03 & \third{61.69} & \second{0.474}& \third{2.92} & 57.42 & 0.309& 3.61 & \best{60.74} & 0.485& 4.16 & \best{69.90} & 0.446& 3.71 & \best{65.52} & \third{0.512} \\
\midrule
PixIE (Ours)& 4.23 & 60.67 & \best{0.478}& \best{2.78} & \third{59.78} & \third{0.333}& 3.77 & 58.46 & \best{0.541}& \second{3.93} & \third{67.70} & \third{0.450}& \third{3.40} & \third{64.36} & \best{0.515} \\

\bottomrule
\end{tabular}}
\label{tab:llie_extra}
\end{table*}

\begin{figure*}[t]
    \centering
     \vspace{-2mm}
    \includegraphics[width=\linewidth]{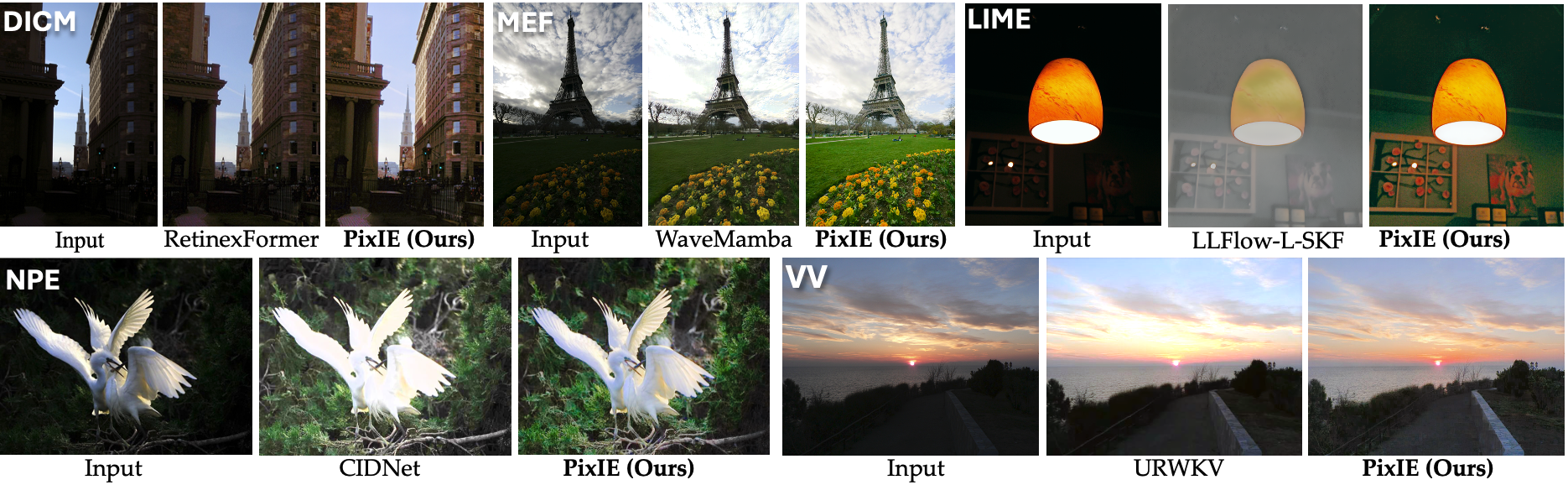}
    \caption{Qualitative comparison of PixIE with recent state-of-the-art methods on five unpaired datasets to show generalization.}
    \label{fig:vis_unpaired}
    \vspace{-4mm}
\end{figure*}

\section{Experiments}
\subsection{Experimental Settings} 
\paragraph{\textbf{Datasets.}} We train and evaluate on LOLv1~\cite{Chen2018Retinex}, LOLv2~\cite{yang2021sparse} and LSRW~\cite{hai2023r2rnet}. LOLv2 contains real-captured and synthetic subsets. LOLv2-Real provides 789 paired images (predefined 689 for training, 100 for testing). We follow the standard train/test splits of 485/15 for LOLv1, 689/100 for LOLv2-Real, and 900/100 for LOLv2-Synthetic.  LSRW is a large-scale real-world paired dataset with 5650 image pairs captured across diverse scenes.  For these paired datasets, we report full-reference distortion metrics PSNR and SSIM, along with the perceptual metric LPIPS~\cite{zhang2018lpips}. For unpaired evaluation, we test on the LIME~\cite{guo2016lime}, NPE~\cite{wang2013naturalness}, MEF~\cite{ma2015perceptual}, DICM~\cite{lee2013contrast}, and VV~\cite{vonikakis2018evaluation} datasets. As there is no ground truth, we use no-reference metrics: NIQE~\cite{mitall2013niqe}, MUSIQ~\cite{ke2021musiq}, and CLIP-IQA~\cite{wang2022clipiqa}.

\paragraph{\textbf{Implementation.}} Our model is implemented in PyTorch and trained with the Adam optimizer~\cite{Kingma2014AdamAM} for $1.2\times10^{5}$ iterations on a GH200 GPU. The learning rate is initialized to $2\times10^{-4}$ and cosine-annealed to $1\times10^{-7}$. During training, images are randomly cropped to $256\times256$ and augmented with random rotations and horizontal/vertical flips. The training objective is 
$\mathcal{L}=\mathcal{L}_{1}+\lambda\,\mathcal{L}_\mathrm{LPIPS}$,
where  $\lambda=0.1$, $\mathcal{L}_{1}$ denotes the $\ell_{1}$ reconstruction loss, and $\mathcal{L}_{\mathrm{LPIPS}}$ is the perceptual loss. 

PixIE has 11.84M trainable parameters (32.84M including frozen DINOv3), uses 63.88 GFLOPs, and runs in 0.081s per image on LOLv2-Real. GFLOPs and inference time include DINOv3 frozen DINOv3 forward pass. Further details are in Supp. Mat. Sec.~A.

\subsection{Performance Comparison}

Quantitative results are shown in ~\cref{tab:llie_main}. PixIE achieves the highest PSNR on three of four datasets and the best LPIPS on three of four. On LOLv1, PixIE achieves the lowest LPIPS, while CIDNet reports a higher PSNR. 
Notably, despite the challenging pixel-misaligned LSRW dataset, PixIE reaches the best PSNR and LPIPS, outperforming WaveMamba and DiffLL. 
On LOLv2-Synthetic, it ranks second in LPIPS while achieving the highest PSNR. On LOLv2-Real, it improves PSNR by 0.22 dB over DiffLL while achieving the highest SSIM (0.902), demonstrating strong reconstruction fidelity and perceptual quality under real low-light conditions.

Qualitative comparisons are shown in~\cref{fig:vis_results}. Although WaveMamba suggests that high-frequency components have limited importance for LLIE, we find fine textures remain crucial under severe low-light noise, where aggressive denoising or illumination correction suppresses structural details. On LOLv1 (top row), most methods fail to recover subtle clothing textures, whereas PixIE preserves these high-frequency patterns. On LOLv2-Real (middle row), competing methods exhibit residual artifacts, over-smoothing, or washed-out colors, struggling to separate the table-leg boundary from the background floor. In contrast, PixIE restores ping-pong table wheels, cleaner edges, and more faithful textures and colors, consistent with its superior LPIPS. On LSRW (bottom row), PixIE preserves fine text (e.g., `SVC-30KVA') with less color bleeding and greater local contrast. As shown in~\cref{fig:LoLv1v2}, PixIE better preserves fine textures and text details while suppressing low-light noise more effectively.

To evaluate generalization, we further test PixIE on five widely used unpaired benchmarks. Quantitative results are reported in~\cref{tab:llie_extra}. 
PixIE achieves the best CLIP-IQA on LIME, DICM and MEF and consistent top-three CLIP-IQA on all five datasets. ~\cref{fig:vis_unpaired} further shows that PixIE balances perceptual quality and distortion removal while preserving more scene details than competing methods. For UHD generalization, we show cropped UHD-LL~\cite{UHDFourICLR2023} results against WaveMamba (UHD-based) and LLFlow-L-SKF (semantic-guided) in~\cref{fig:teaser} (left); PixIE denoises better while preserving details. We further evaluate object detection on ExDark~\cite{Exdark}. PixIE achieves the best mean AP50, showing downstream benefit. Additional qualitative results, failure case analysis, and evaluation on low-light videos are provided in Supp. Mat. Sec.~B--D.


\begin{table}[t]
\vspace{-1mm}
\caption{Per-category object detection mean AP (\%) on ExDark. Cells highlighted in \best{red}, \second{orange}, and \third{yellow} denote the best, second-best, and third-best results, respectively.}
\centering
\scriptsize
\setlength{\tabcolsep}{2.2pt}
\resizebox{\columnwidth}{!}{%
\begin{tabular}{@{}lcccccccccccc|c@{}}
\toprule
Method & Bic. & Boat & Bot. & Bus & Car & Cat & Chr. & Cup & Dog & Mot. & Ppl. & Tbl. & Mean \\
\midrule
ZeroDCE  &75.8 &66.5 &65.6 &84.9 &77.2 &56.3 &53.8 &59.0 &63.5 &64.0 &68.3 &\second{46.3} &65.1  \\
EnlightenGAN  &79.6  & 73.0 &71.4  &85.6  & \third{80.4} & 60.5 &64.1  &69.4  &72.7  &63.2  &77.5  &43.2  & 70.1 \\
Restormer  &76.2 &65.1 &64.2 &84.0 &76.3 &59.2 &53.0 &58.7 &66.1 &62.9 &68.6 &45.0 &64.9 \\
Retinexformer & 76.3 &66.7 &65.9 &84.7 &77.6 &61.2 &53.5 &60.7 &67.5 &63.4 &69.5 &\third{46.0} &66.1 \\
DiffLL  &81.6  &71.4  & 67.7 & 83.9 & 77.1 & 57.7 & 62.4 & 64.4 & 70.3 & 60.6 & 73.1& 44.2& 67.8\\
LLFlow-L-SKF  & 80.6 & 71.7 & 70.3 & 82.1 & 78.6 & 60.9 & 63.0 & \third{70.2} & 74.8 & 61.7 & 77.2 & \best{48.1} & 69.9 \\
WaveMamba  & \third{81.9}  & \best{78.2} & 70.5 & \third{86.5} & \second{80.6} & \second{64.9} & \second{67.2} & \best{73.6} & \second{77.5} & \best{66.4} & 79.1 & 41.8 & \second{72.3} \\
URWKV &79.5 &\second{77.2} &\second{72.2} &\second{86.7} &\best{81.0} &\third{61.7} &\third{65.5} &68.3 &74.0 &\third{64.7} &\second{78.6} &41.6 &\third{70.9} \\
CIDNet & \second{82.8} & 75.4 & \third{71.8} & 84.7 &80.1 &59.9 &64.5  &69.2  &\third{75.7} & 64.3 & \third{78.4} &41.0  & 70.7\\ \hline
\textbf{PixIE (Ours)} & \best{82.9} & \third{76.5} & \best{72.9} & \best{86.9} & 79.7 & \best{65.9} & \best{67.3} & \second{71.1} & \best{79.4} & \second{65.0} & \best{80.0} & 45.5 & \best{\textbf{72.8}} \\
\bottomrule
\end{tabular}%
}
\vspace{-4mm}
\label{tab:per_category}
\end{table}

\subsection{Ablation Study}
\label{sec:abl}

~\cref{tab:ablation} shows that removing Cross-scale Denoising by feeding raw features ($F_0$) into MRPE reduces PSNR by $1.54$\,dB and increases LPIPS from $0.089$ to $0.181$. Applying PPBs before denoising also lowers PSNR, highlighting the importance of denoising before prompting. Replacing MRPE with a point-wise $1{\times}1$ embedding reduces PSNR by $0.69$\,dB, while single $3{\times}3$, $5{\times}5$ convolutions and an equal channel split are also less effective, supporting multi-receptive-field aggregation at the embedding stage. Removing PPB reduces PSNR from $29.08$ to $27.50$\,dB, while removing FM prompting (DINOv3) lowers it to $27.94$\,dB, demonstrating that FM guidance further enhances PPB.

\begin{table}[t]
\vspace{-1mm}
\caption{Component ablation on LOLv2-Real. PixIE uses a 25/37.5/37.5 MRPE split and DINOv3 layers $[2,5,8,11]$.}
\label{tab:ablation}
\centering
\footnotesize
\setlength{\tabcolsep}{3pt}
\renewcommand{\arraystretch}{1.0}
\begin{tabular*}{\columnwidth}{@{\extracolsep{\fill}}llccc@{}}
\toprule
Module & Variant & PSNR$\uparrow$ & SSIM$\uparrow$ & LPIPS$\downarrow$ \\
\midrule
\multirow{5}{*}{Components} & w/o Cross-scale Denoising  & 27.54 & 0.850 & 0.181 \\
 & PPB before Denoising        & 28.38	& 0.894	&0.112 \\
 & w/o MRPE ($1\times1$ only) & 28.39 & 0.893 & 0.108 \\
 & w/o PPB                    & 27.50 & 0.887 & 0.114 \\
 & w PPB w/o FM                 & 27.94 & 0.891 & 0.112 \\
\midrule
\multirow{3}{*}{MRPE}
 & $3\times3$ conv            & 27.98 & 0.894 & 0.107 \\
 & $5\times5$ conv            & 28.07 &0.895 &0.110    \\
 & Equal split                & 27.51 & 0.892 & 0.111 \\
\midrule
\multirow{4}{*}{FM in PPB}
 & SAM2                       & 28.10 & 0.898 & 0.100 \\
 & CLIP                       & 27.26 & 0.890 & 0.107 \\
 & DA3                        & 28.48 & 0.897 & 0.103 \\
 & DINOv3 (shuffled tokens)   & 27.59 &0.894	&0.103 \\
\midrule
\multirow{3}{*}{DINOv3 layers}
 & $[11]$                     & 27.38 & 0.894 & 0.100 \\
 & $[8,9,10,11]$              & 27.98 & 0.895 & 0.098 \\
 & $[0,1,2,3]$                & 28.40 & 0.899 & 0.105 \\
\midrule
\multicolumn{2}{@{}l}{PixIE (DINOv3)} & 29.08 & 0.902 & 0.089 \\
\bottomrule
\end{tabular*}
\vspace{-2mm}
\end{table}

PPB can incorporate different foundation models. SAM2 and DA3 (Depth Anything v3) improve over PPB without FM prompting, while CLIP slightly reduces PSNR despite improving LPIPS. DINOv3 achieves the best overall performance and is therefore adopted throughout. Spatially shuffling its tokens reduces PSNR to $27.59$\,dB, below PPB without FM ($27.94$\,dB), indicating that spatial correspondence is important for effective prompting. For layer selection, early layers $[0,1,2,3]$ outperform late layers $[8,9,10,11]$ by $0.42$\,dB, with the last layer $[11]$ alone achieving lower PSNR, while sampling across depth $[2,5,8,11]$ gives the best results. Further FM configurations and comparisons of feature robustness under low-light are provided in Supp. Mat. Sec.~E.

\begin{table}[t]
\caption{Ablation of conditioning methods on LOLv2-Real with DINOv3 prompts. Params (trainable, excluding DINOv3) and GFLOPs ($\downarrow$) are for the full model on $256\times256$ inputs; GFLOPs include the frozen DINOv3 encoder. 
}
\label{tab:abl_conditioning}
\centering
\footnotesize
\setlength{\tabcolsep}{2pt}
\renewcommand{\arraystretch}{1.0}
\begin{tabular*}{\columnwidth}{@{\extracolsep{\fill}}lccccc@{}}
\toprule
Variant & PSNR$\uparrow$ & SSIM$\uparrow$ & LPIPS$\downarrow$ & Params & GFLOPs \\
\midrule
Cross-attn                       & 27.84 & 0.898 & 0.098 &10.58 &74.43 \\
Global broadcast                      & 28.05 & 0.890 & 0.097 &11.79 &61.33\\
Patch-wise                & 26.98 & 0.764 & 0.151 & 11.79& 	61.77\\
Per-pixel (PixelDiT~\cite{yu2025pixeldit})   & 27.81 & 0.891 &0.113& 539.58  & 128.97 \\
Spatially adaptive & 28.36 & 0.898 & 0.108 &11.84 &131.62 \\
\midrule
Ours (SCMo)         & 29.08 & 0.902 & 0.089 & 11.84 &63.88 \\
\bottomrule
\end{tabular*}
\end{table}

\begin{figure}[t]
    \centering
    \vspace{-2mm}
    \includegraphics[width=\linewidth]{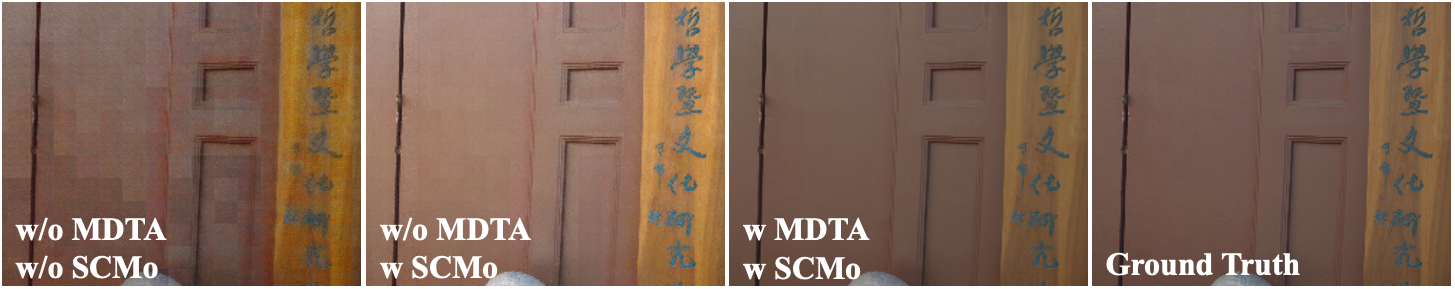}
    \vspace{-2mm}
    \caption{Qualitative ablation of PixIE on the LOLv2-Real. We compare (i) w/o MDTA and w/o SCMo, (ii) w/o MDTA and w SCMo, and (iii) w MDTA and w SCMo (ours).}
    \label{fig:abl}
    \vspace{-4mm}
\end{figure}

~\cref{tab:abl_conditioning} shows that SCMo outperforms other conditioning methods, including spatially adaptive modulation~\cite{wang2018sftgan}, which upsamples tokens before predicting per-pixel parameters, and per-pixel modulation~\cite{yu2025pixeldit}, at roughly half the GFLOPs. The latter requires $539.58$M trainable parameters due to dense $P^2$ prediction.
The visual comparisons in~\cref{fig:abl} further illustrate the role of SCMo and MDTA. Patch-wise conditioning (panel~(i)) produces visible boundary seams, which SCMo substantially reduces even without MDTA (panel~(ii)). Adding MDTA further reduces spatial inconsistencies and improves fine textures by enabling non-local feature interactions (panel~(iii)). 

~\cref{tab:comp_ablation} evaluates SCC. We set $P=16$ to match the DINOv3 patch size, aligning the token grid with conditioning features without resampling. Spatial folding alone expands channels to $CP^2$ and increases parameters to 48.90M, while compression alone reduces PSNR compared with No SCC. Combining both achieves the highest PSNR at the lowest computational cost. Increasing $D_{\mathrm{attn}}$ from 16 to 32 or 64 adds parameters and GFLOPs without improving performance. Replacing MDTA with spatial multi-head self-attention (MHSA) reduces PSNR by $0.95$,dB at comparable complexity. Spatial attention over uncompacted pixel tokens is impractical, as its cost grows quadratically with $HW$~\cite{Zamir2021Restormer}. We thus adopt $D_{\mathrm{attn}}=16$ with MDTA.

\begin{table}[t]
\caption{Ablation of Spatial-Channel Compaction (SCC) on LOLv2-Real. $P$: spatial fold size ($P{=}1$: no folding; $P{=}16$ matches the DINOv3 patch size). $D_{\mathrm{attn}}$: attention dimension ($C$ or $CP^2$: no compression; $C$: channel width). 
Params (M, trainable, excluding DINOv3) and GFLOPs ($\downarrow$) are for the full model on $256\times256$ inputs; GFLOPs include the frozen DINOv3 encoder.
}
\label{tab:comp_ablation}
\centering
\footnotesize
\setlength{\tabcolsep}{2pt}
\renewcommand{\arraystretch}{1.0}
\begin{tabular*}{\columnwidth}{@{\extracolsep{\fill}}lcccccrr@{}}
\toprule
Variant & $P$ & $D_{\mathrm{attn}}$ & PSNR$\uparrow$ & SSIM$\uparrow$ & LPIPS$\downarrow$ & Params & GFLOPs \\
\midrule
\multicolumn{8}{@{}l}{\textbf{\textit{Compaction type}}} \\
No SCC       & 1  & $C$    & 28.38 & 0.897 & 0.105 & 4.80  & 75.54 \\
Channel only & 1  & 16     & 27.82 & 0.896 & 0.102 & 4.30  & 66.24 \\
Spatial only & 16 & $CP^2$ & 27.85 & 0.895 & 0.106 & 48.90 & 67.40 \\
\midrule
\multicolumn{8}{@{}l}{\textbf{\textit{Compaction dimension}} (full SCC)} \\
& 16 & 64 & 28.08 & 0.879 & 0.103 & 34.07 & 66.75 \\
                           & 16 & 32 & 28.22 & 0.897 & 0.098 & 19.23 & 64.83 \\
\midrule
\multicolumn{8}{@{}l}{\textbf{\textit{Attention on compact tokens}}} \\
Spatial MHSA & 16 & 16 &28.13 &0.896 &0.114 &11.84 &63.90 \\
\midrule
Ours (MDTA) & 16 & 16 & 29.08 & 0.902 & 0.089 & 11.84 & 63.88 \\
\bottomrule
\end{tabular*}
\end{table}

\section{Conclusion}
We introduced PixIE, a feed-forward pixel-space framework for low-light image enhancement that integrates foundation-model priors through spatially continuous modulation (SCMo). PixIE combines cross-scale denoising with FM-guided pixel refinement, supported by Multi-Receptive-Field Pixel Embedding (MRPE) for neighborhood-aware feature extraction and Spatial-Channel Compaction (SCC) for efficient multi-scale pixel-space attention. Experiments on standard LLIE benchmarks demonstrate strong quantitative and perceptual performance while preserving fine textures in practical scenarios.

\section*{Acknowledgments}
This work was supported by the UKRI MyWorld Strength in Places Programme (SIPF00006/1).

{
    \small
    \bibliographystyle{ieeenat_fullname}
    \bibliography{main}
}


\end{document}


\makeatletter
\g@addto@macro\@maketitle{%
    \begin{center}
        \vspace{-2mm}
        \includegraphics[width=\textwidth, trim=0 2cm 0 0, clip]{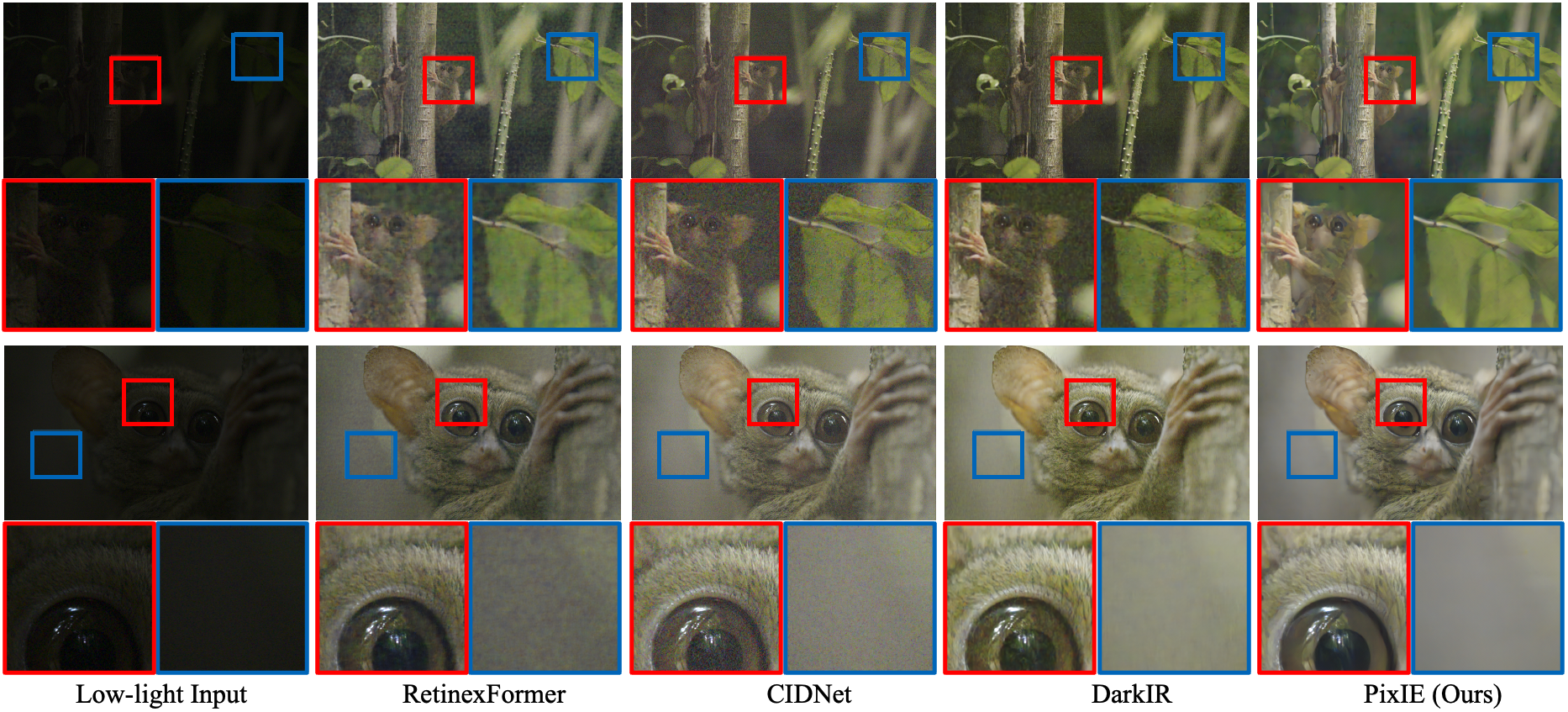}
        \includegraphics[width=\textwidth]{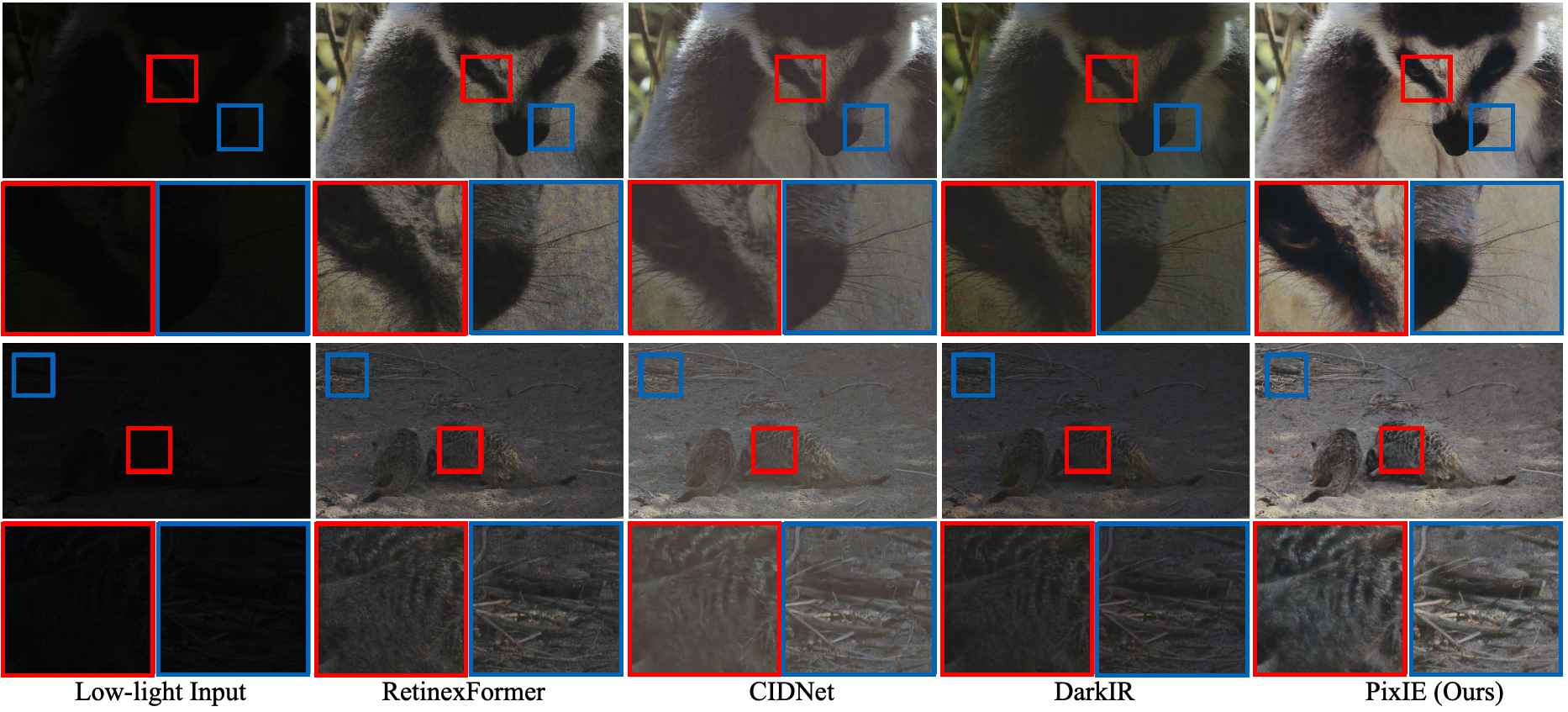}
        \captionof{figure}{Qualitative results on real, unseen, professionally filmed low-light footage.  PixIE effectively suppresses noise while preserving fine textures in fur, leaves, and ground details, outperforming competing methods.}
        \label{fig:vis_real_tarsier}
        \vspace{4mm}
    \end{center}
}
\makeatother

\maketitle

\appendix

\section{Complexity Comparisons}

\begin{table}[t]
\centering
\caption{Model complexity and performance on LOLv2-Real.}
\label{tab:complexity_lolv2real}
\small
\resizebox{\columnwidth}{!}{%
\begin{tabular}{l|cccc}
\toprule
Method & Params (M)$\downarrow$ & GFLOPs$\downarrow$ & Inference Time (s)$\downarrow$ & PSNR$\uparrow$ \\
\midrule
ZeroDCE~\cite{Zero-DCE} & 0.075 &4.83 & 0.005& 18.06\\
EnlightenGAN~\cite{jiang2021enlightengan}  &114.35 & 61.01  &0.011 &18.64 \\
Restormer~\cite{Zamir2021Restormer} & 26.13 & 144.25 & 0.513 &18.60 \\
RetinexFormer~\cite{Cai_2023_ICCV} & 1.61  & 15.57 &0.046& 22.79 \\
DiffLL~\cite{jiang2023low} & 22.08 & 89.79 & 0.157 &28.86 \\
LLFlow-L-SKF~\cite{wu2023skf} & 17.43 &358.4 &0.130 &26.20 \\
FourierDiff~\cite{lv2024fourier} &547.56 &4277.52 & 9.882 & 16.85 \\
WaveMamba~\cite{zou2024wavemamba}  &  1.26 & 1.44 & 0.056 & 22.13 \\
MambaLLIE~\cite{MambaLLIE} &2.28 &20.85 & 0.073& 22.95\\
URWKV~\cite{xu2025urwkv} & 2.25 &18.34 &0.117 & 23.11 \\
CIDNet~\cite{Yan_HVICID}           &  1.88 & 7.57  & 0.058 & 24.11 \\

\midrule
PixIE (Ours)                       & 11.84 & 63.88 & 0.081& 29.08 \\
\bottomrule
\end{tabular}}
\vspace{-1pt}
\end{table}

\noindent
We analyze PixIE's computational cost and performance trade-offs against state-of-the-art low-light enhancement methods. Reported GFLOPs and inference time include the DINOv3 forward pass, while PixIE's 11.84\,M trainable parameters exclude the frozen DINOv3 ViT-S/16 (21\,M), giving 32.84\,M parameters in total. 

As shown in Table~\ref{tab:complexity_lolv2real} and Fig.~1 (right) of the main paper, PixIE lies on the upper-left PSNR--GFLOPs Pareto frontier. Compared with lightweight CIDNet~\cite{Yan_HVICID} and WaveMamba~\cite{zou2024wavemamba}, PixIE achieves nearly $5$\,dB higher PSNR at greater computational cost. It also improves over Restormer~\cite{Zamir2021Restormer} by 10.48\,dB using only 44\% of its GFLOPs, and over semantic-guided LLFlow-L-SKF~\cite{wu2023skf} by 2.88\,dB with 5.6$\times$ lower computational cost, demonstrating a favorable fidelity--efficiency trade-off. We profile standard operators using fvcore and analytically account for custom CUDA selective-scan operations in Mamba-based models (\eg WaveMamba). For diffusion methods, FLOPs include all inference steps (10 for DiffLL~\cite{jiang2023low} and 100 for FourierDiff~\cite{lv2024fourier}). FourierDiff uses a frozen ${\sim}552$M-parameter guided-diffusion ImageNet backbone at $256\times256$, explaining its high parameter count and, together with 100 sampling steps, its computational cost.

\begin{table*}[!ht]
\centering
\caption{Additional quantitative comparisons on LOLv1, LOLv2-Real and LOLv2-Synthetic.}
\small
\setlength{\tabcolsep}{4pt}
\setlength{\arrayrulewidth}{0.4pt}
\vspace{-1mm}
\begin{tabular}{lccc|ccc|ccc}
\toprule
\multirow{2}{*}{Method}
& \multicolumn{3}{c|}{LOLv1}
& \multicolumn{3}{c|}{LOLv2-Real}
& \multicolumn{3}{c}{LOLv2-Synthetic} \\
\cmidrule(lr){2-4} \cmidrule(lr){5-7} \cmidrule(l){8-10}
& PSNR$\uparrow$ & SSIM$\uparrow$ & LPIPS$\downarrow$
& PSNR$\uparrow$ & SSIM$\uparrow$ & LPIPS$\downarrow$
& PSNR$\uparrow$ & SSIM$\uparrow$ & LPIPS$\downarrow$ \\
\midrule
DarkIR~\cite{Feijoo_2025_CVPR} (CVPR'25)
& \second{26.19} & \third{0.865} & 0.163
& \third{23.87} & \second{0.880} & 0.192
& 25.54 & 0.934 & 0.053 \\
Diff-Retinex++~\cite{yi2025diff} (TPAMI'25)
& 24.64 & \second{0.866} & \third{0.101}
& 23.13 & \third{0.870} & \second{0.136}
& \second{26.01} & \second{0.944} & \best{0.040} \\

Multinex~\cite{multinex2026} (CVPR'26)
& 23.19 & 0.843 & 0.129
& 23.04 & 0.860 & 0.178
& 25.04 & 0.930 & 0.068 \\
InterLight~\cite{wang2026interlight} (IJCAI'26)
& \third{24.78} & 0.862  & \second{0.098}
& \second{24.06}  & 0.866  & \second{0.136}
& \third{25.73} &\third{0.935} & \third{0.050} \\
\midrule
\textbf{PixIE (Ours)}
& \best{27.07} & \best{0.876} & \best{0.073}
& \best{29.08} & \best{0.902} & \best{0.089}
& \best{29.10} & \best{0.950} & \second{0.042} \\
\bottomrule
\end{tabular}%
\label{tab:supp_lol_comparison}
\vspace{1mm}
\end{table*}

\begin{figure*}[!ht]
    \centering
    \vspace{-1mm}
    \includegraphics[width=\linewidth]{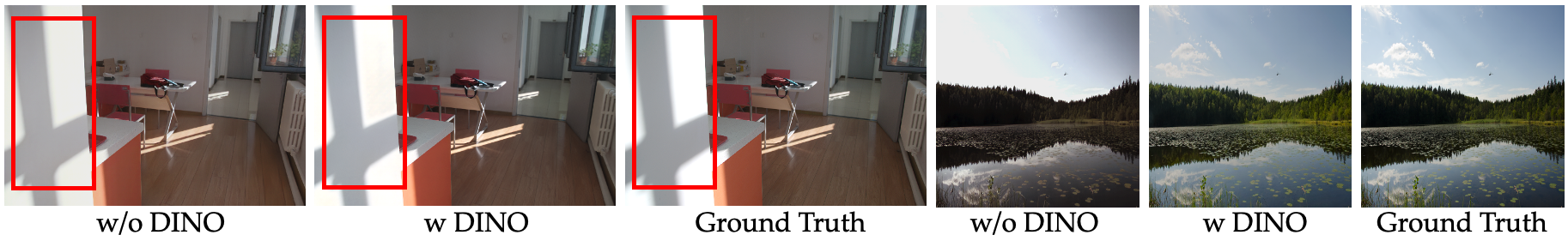}
    \caption{Qualitative comparison with and without DINO semantic conditioning in the Prompted Pixel Block (PPB). The left half shows results on LOL-v2 Real, and the right half shows results on LOL-v2 Synthetic. DINO conditioning improves color balance and brightness restoration, yielding results closer to the ground truth.}
    \label{fig:DINO_abla_vis}
    \vspace{-1mm}
\end{figure*}

\section{Additional Evaluation}

\subsection{Real Low-light Visual Results}

As shown in \cref{fig:vis_real_tarsier}, PixIE more effectively suppresses background noise around the animal and leaves in the first row, while removing noise from the eye without sacrificing fine fur textures in the second. The bottom two rows further demonstrate improved noise suppression and preservation of fine fur and ground textures. Notably, DarkIR~\cite{Feijoo_2025_CVPR} is trained to handle compound degradations, including blur, whereas PixIE, CIDNet, and RetinexFormer are trained on LOLv2-Real. These previously unseen real low-light scenes lack ground-truth references, providing qualitative evidence of real-world generalization.


\subsection{Additional Quantitative Results}

We further compare PixIE with other recent low-light restoration methods on LOLv1, LOLv2-Real, and LOLv2-Synthetic, including DarkIR~\cite{Feijoo_2025_CVPR}, which jointly addresses low-light enhancement, denoising, and deblurring. As shown in \cref{tab:supp_lol_comparison}, PixIE achieves the best PSNR and SSIM across all three datasets, as well as the lowest LPIPS on LOLv1 and LOLv2-Real.

\subsection{Effect of DINO Semantic Conditioning}

\noindent
Fig.~\ref{fig:DINO_abla_vis} shows qualitative results with and without DINO semantic prompts. Without DINO features, the model exhibits a clear color shift. Adding DINO conditioning improves both color balance and scene brightness, producing results that more closely match the ground truth. This indicates that DINO semantic features provide a global semantic context that benefits both color and illumination restoration. 

\section{Failure Case Analysis} 
\begin{figure*}
    \centering
    \vspace{-0.1mm}
    \includegraphics[width=\linewidth]{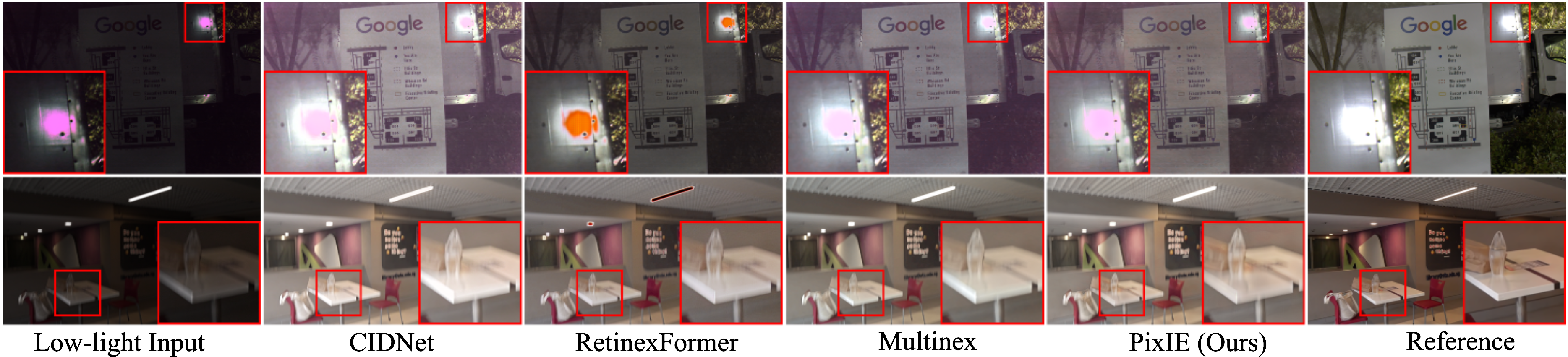}
    \caption{Failure cases under challenging degradations. Top: Spatially non-uniform color casts from artificial lighting. Bottom: Incomplete recovery of fine details under low-light with severe blur.}
    \label{fig:fail}
    \vspace{-2mm}
\end{figure*}

\noindent We identify three limitations: (i) extreme spatially non-uniform color casts from artificial lighting, which remain challenging across methods and can lead to inaccurate color (\cref{fig:fail}, top). RetinexFormer largely removes the pink cast but introduces pronounced color artifacts around artificial lights; (ii) very small objects ($<$16\,px), which fall below the DINO patch resolution and receive weak semantic guidance; and (iii) compound degradations involving low light, noise, and blur~\cite{Feijoo_2025_CVPR}, which PixIE is not explicitly trained to handle. Although partially robust, PixIE struggles to recover details under severe blur (\cref{fig:fail}, bottom). 
Addressing these limitations remains future work.

\begin{table*}[!htb]
\centering
\caption{Quantitative comparison on the BVI-RLV and DID datasets.
Higher PSNR and SSIM indicate better performance, while lower LPIPS
and tLPIPS indicate better perceptual quality and temporal consistency,
respectively.}
\small


\begin{tabular}{l|cccc|cccc}
\toprule
\multirow{2}{*}{Methods}
& \multicolumn{4}{c|}{BVI-RLV~\cite{lin2024bvirlv}}
& \multicolumn{4}{c}{DID~\cite{Fu_2023_ICCV}} \\
\cmidrule(lr){2-5}
\cmidrule(lr){6-9}
& PSNR$\uparrow$
& SSIM$\uparrow$
& LPIPS$\downarrow$
& tLPIPS$\downarrow$
& PSNR$\uparrow$
& SSIM$\uparrow$
& LPIPS$\downarrow$
& tLPIPS$\downarrow$ \\
\midrule
BVI-Mamba~\cite{huang2025bvi} 
        & 20.47 & 0.816 & \third{0.157} & 0.160
        & \best{31.22} & \best{0.912} & \best{0.071} & \third{0.124}\\

LLVE-STCD~\cite{xu2025low} & \best{33.43} & \second{0.919} & \second{0.104}  & \second{0.034}
        &26.93 & \third{0.898} & \third{0.141} & \second{0.015} \\

ReMARF~\cite{liu2025rmfat} & \third{24.26} & \third{0.852} & 0.231 & \third{0.148} & \second{30.46} & 0.869 & 0.175 & - \\

        
\midrule
PixIE (Ours) & \second{32.80} & \best{0.937} & \best{0.083} & \best{0.006} & \third{28.95} & \best{0.912} & \second{0.084} & \best{0.011} \\

\bottomrule
\end{tabular}%
\label{tab:video_results}
\end{table*}

\section{LLVE Evaluation}
\noindent 
We further evaluate PixIE on low-light video enhancement (LLVE) benchmarks to assess its applicability beyond single-image restoration. 
We train and evaluate PixIE frame by frame, without temporal modeling or video-specific modifications, and compare it with recent LLVE methods. As shown in \cref{tab:video_results}, PixIE achieves competitive PSNR and the best LPIPS and tLPIPS on BVI-RLV, while achieving the best tLPIPS and jointly best SSIM on DID, despite lacking explicit temporal modeling.

\section{Foundation Model Comparisons}

\begin{figure}[ht]
    \centering
    \vspace{-1mm}
    \includegraphics[width=\columnwidth]{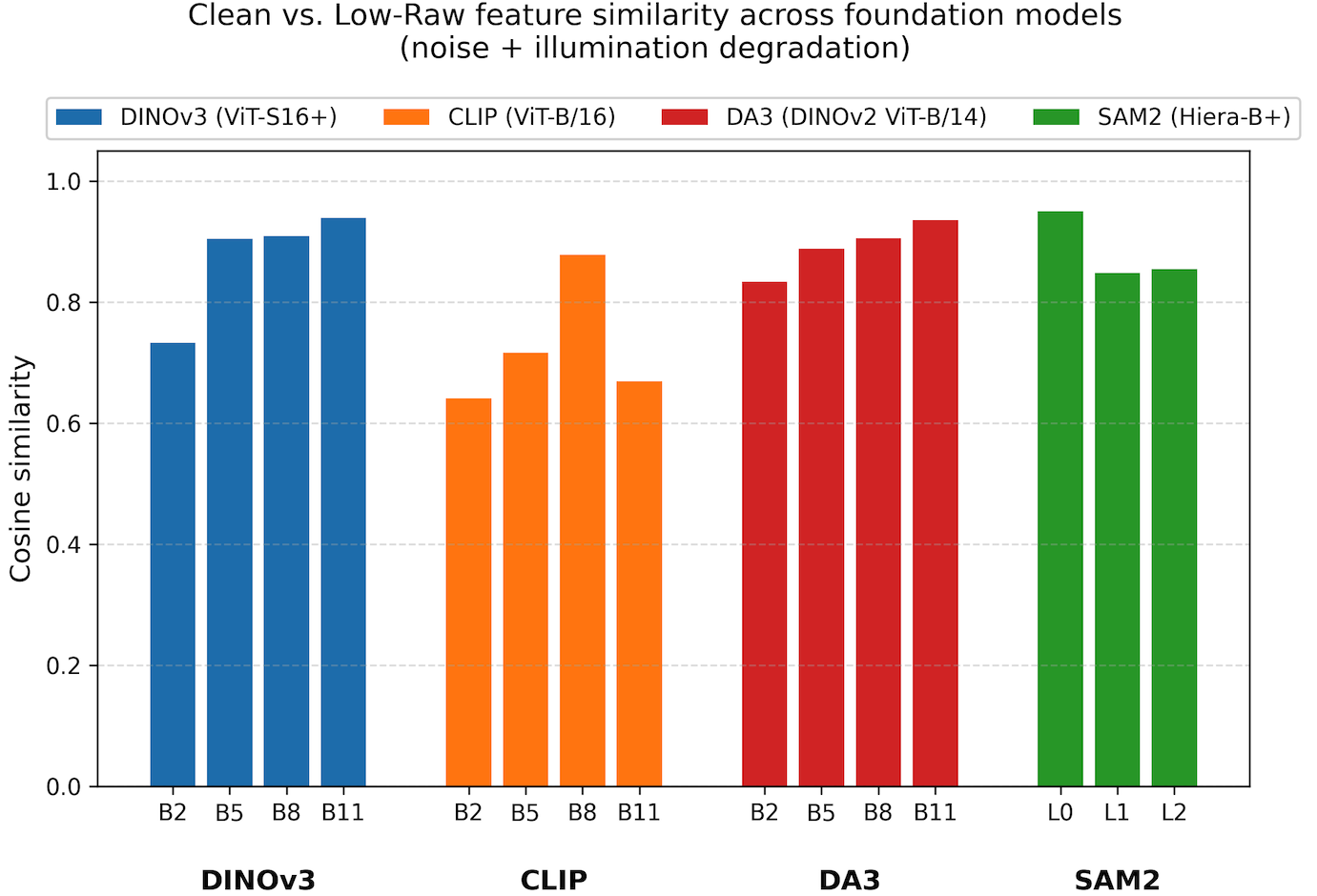}
    \caption{Feature similarity between clean and degraded low-light images across different foundation models. Cosine similarities are computed within each model's feature space; therefore, only degradation trends within each model should be compared.}
    \label{fig:fm_similarity}
    \vspace{-1mm}
\end{figure}

To further motivate the choice of DINOv3, Fig.~\ref{fig:fm_similarity} compares feature similarity between clean and degraded low-light images across different foundation models. Since different FMs produce features in different embedding spaces, absolute cosine values are not directly comparable across models. Instead, we compare the degradation trend within each model. DINOv3 exhibits the smallest degradation at the deepest selected layer (B11). We attribute its superior restoration performance to its general-purpose semantic representations for recognition and scene understanding, whereas DA3 is primarily optimized for monocular depth estimation.

\paragraph{DINOv3.}
We use the frozen DINOv3 ViT-S/16Plus encoder ($D_d{=}384$) and extract intermediate patch tokens from layers $\{2,5,8,11\}$ to prompt the four PPBs at each scale.

\paragraph{CLIP.}
We use the frozen OpenCLIP ViT-B/16 encoder ($D_d{=}768$) with the same layer selection. Following the official implementation, inputs are resized to $224\times224$, producing a $14\times14$ token grid and consequently spatially coarser conditioning. Table~\ref{tab:CLIPabl_conditioning} shows conditioning mechanisms using CLIP with PPBs. This suggests that the effectiveness of the conditioning mechanism depends on the semantic quality and spatial specificity of the underlying foundation-model representations. This further motivates our choice of DINOv3, which provides the strongest overall restoration performance among the evaluated foundation models.

\begin{table}[!htb]
\vspace{-1mm}
\centering
\footnotesize
\setlength{\tabcolsep}{3pt}
\renewcommand{\arraystretch}{1.0}
\caption{Ablation of conditioning mechanisms using CLIP prompts on
LOLv2-Real dataset.}

\begin{tabular*}{\columnwidth}{@{\extracolsep{\fill}}lccc@{}}
\toprule
Method & PSNR$\uparrow$ & SSIM$\uparrow$ & LPIPS$\downarrow$ \\
\midrule
Cross-attn fusion       & 27.40 & 0.888 & 0.102  \\
Global broadcast        & 26.39 & 0.882 & 0.120 \\
Patch-wise conditioning & 26.69  & 0.888 & 0.119 \\
Per-pixel prediction    & 27.15 &0.889	&0.111   \\
\midrule
Ours (SCMo)             & 27.26 & 0.890 & 0.107 \\
\bottomrule
\end{tabular*}
\label{tab:CLIPabl_conditioning}
\vspace{-3mm}
\end{table}

\paragraph{Depth Anything 3.}
We use the frozen Depth Anything 3 (DA3-BASE) DINOv2 ViT-B/14 encoder ($D_d{=}768$), retaining only its backbone and discarding the DPT depth head and camera encoder-decoder. Intermediate patch tokens from the same selected layers are used to prompt the four PPBs at each scale in the monocular (no camera-token) setting.

\paragraph{SAM2.}
We use the frozen SAM2.1 Hiera-B+ image encoder. Three FPN feature levels are projected to a common feature dimension ($D_d{=}256$) via $1\times1$ convolutions and used to prompt the PPBs.

\noindent
These variants differ in feature dimensions and native feature representations and are therefore not intended as a strictly controlled comparison between FMs. This motivates choice of DINOv3, which provides the strongest overall restoration performance among evaluated foundation models.

\section{More Ablation Studies}

\paragraph{\textbf{Modulation Field Upsampling Strategy.}}
Table~\ref{tab:ablation_upsampling} compares strategies for upsampling the modulation field to the full resolution. PixelShuffle runs out of memory because it requires large full-resolution intermediate feature maps. Bicubic interpolation is more memory-efficient than PixelShuffle. Our design uses bilinear upsampling, which provides the best trade-off between restoration quality and efficiency among the evaluated strategies. It avoids the memory overhead of learned upsampling while achieving the best results, with a low parameter count and competitive GFLOPs.

\begin{table}[ht]
\centering
\caption{Ablation on modulation field upsampling strategy. Metrics are reported on the LOLv2-Real dataset with complexity measured for $256\times256$ inputs.}
\label{tab:ablation_upsampling}
\setlength{\tabcolsep}{3.5pt}
\renewcommand{\arraystretch}{1.1}
\footnotesize
\resizebox{\columnwidth}{!}{%
\begin{tabular}{lccc}
\toprule
Strategy & PSNR$\uparrow$/SSIM$\uparrow$/LPIPS$\downarrow$ & Params (M)$\downarrow$ & GFLOPs$\downarrow$ \\
\midrule
PixelShuffle & OOM  &122.37 & 805.94 \\
Bicubic &  28.22/0.897/0.103 & 11.84 & 64.38 \\
\midrule
Bilinear (PixIE) & 29.08/0.902/0.089 & 11.84 & 63.88 \\
\bottomrule
\end{tabular}}
\end{table}

\section{Transformer Block Details}
In the Cross-Scale Denoising module, we use Restormer-style transformer blocks, each consisting of a Multi-Dconv Head Transposed Attention (MDTA) module followed by a Gated-Dconv Feed-Forward Network (GDFN). Details of each component are given below.
\paragraph{\textbf{Multi-Dconv Head Transposed Attention (MDTA).}}
Given an input feature map $X\in\mathbb{R}^{B\times D\times H\times W}$, we first apply normalization $Y=\mathrm{Norm}(X)$ and form query, key, and value projections using a point-wise convolution followed by a depth-wise convolution:
\begin{equation}
Q = W^{Q}_{d} W^{Q}_{p} Y,\quad
K = W^{K}_{d} W^{K}_{p} Y,\quad
V = W^{V}_{d} W^{V}_{p} Y,
\end{equation}
where $W^{(\cdot)}_{p}$ and $W^{(\cdot)}_{d}$ denote $1\times1$ point-wise and $3\times3$ depth-wise convolutions, respectively. 
MDTA then reshapes $Q,K,V$ such that attention is computed in the \emph{channel domain}, producing a transposed attention map of size $D\times D$:
\begin{align}
A &= \mathrm{Softmax}\!\left(\frac{\hat{K}\hat{Q}}{\alpha}\right), \\
\mathrm{MDTA}(X) &= W_{o}\big(\hat{V}A\big),
\end{align}
where $\hat{Q}\in\mathbb{R}^{HW\times D}$, $\hat{K}\in\mathbb{R}^{D\times HW}$, and $\hat{V}\in\mathbb{R}^{HW\times D}$ are reshaped tensors, $\alpha$ is a learnable scaling parameter, and $W_o$ is a $1\times1$ output projection. We use a multi-head variant by splitting channels into $h$ heads.

\paragraph{\textbf{Gated-Dconv Feed-Forward Network (GDFN).}}
Following MDTA, a gated feed-forward transformation is employed that combines channel mixing with local spatial aggregation:
\begin{equation}
U_i = W^{i}_{d}W^{i}_{p}\,\mathrm{Norm}(X), \qquad i\in\{1,2\},
\end{equation}
\begin{equation}
\mathrm{GDFN}(X)=W^{0}_{p}\!\left(\phi(U_1)\odot U_2\right).
\end{equation}
where $\odot$ denotes element-wise multiplication, $\phi(\cdot)$ is GELU, and
$W_p^{(\cdot)}$ and $W_d^{(\cdot)}$ denote point-wise and depth-wise convolutions.
Each Restormer block is applied in a residual form:
\begin{equation}
X \leftarrow X + \mathrm{MDTA}(X), \qquad
X \leftarrow X + \mathrm{GDFN}(X).
\end{equation}

{
    \small
    \bibliographystyle{ieeenat_fullname}
    \bibliography{main}
}